\documentclass[letterpaper, 10pt, journal, twoside]{IEEEtran}

\IEEEoverridecommandlockouts                              

\usepackage{amsmath} 
\usepackage[colorlinks=true, linkcolor=blue, urlcolor=blue, citecolor=blue]{hyperref}
\usepackage{cleveref}
\usepackage[ruled,vlined,linesnumbered]{algorithm2e}
\usepackage{graphicx}
\usepackage{booktabs}
\usepackage{multirow}
\usepackage{multicol}
\usepackage{subcaption}
\usepackage{subscript}

\usepackage{enumitem}
\usepackage{amsmath}      
\usepackage{amssymb}      
\usepackage{mathpartir}   
\usepackage{makecell}
\usepackage{bm}
\usepackage{pifont}
\usepackage[markup=colored, addedcolor=blue]{changes}

\newcounter{RNum}

\newcommand{\fref}[1]{Fig.~\ref{#1}}
\newcommand{\sref}[1]{Section~\ref{#1}}
\newcommand{\tref}[1]{Table~\ref{#1}}

\newcommand{\myparagraph}[1]{\noindent\textbf{#1}~}
\newcommand{\eref}[1]{(\ref{#1})}

\newcommand{\shortname}{Flax\xspace}

\let\llncssubparagraph\subparagraph
\let\subparagraph\paragraph
\usepackage[compact]{titlesec}
\let\subparagraph\llncssubparagraph
\titlespacing{\section}{0pt}{4pt}{1pt}
\titlespacing{\subsection}{0pt}{3pt}{1pt}
\titlespacing{\subsubsection}{0pt}{2pt}{1pt}
\titlespacing{\paragraph}{0ex}{0pt}{4pt}
\titlespacing{\subparagraph}{0ex}{0pt}{4pt}

\title{\LARGE \bf
Fast Task Planning with Neuro-Symbolic Relaxation
}

\author{Qiwei Du\textsuperscript{1}, Bowen Li\textsuperscript{2}, Yi Du\textsuperscript{1}, Shaoshu Su\textsuperscript{1}, Taimeng Fu\textsuperscript{1}, Zitong Zhan\textsuperscript{1}, Zhipeng Zhao\textsuperscript{1}, and Chen Wang\textsuperscript{1}
\thanks{Manuscript received: July, 21, 2025; Revised December, 4, 2025; Accepted Janurary, 31, 2026. This paper was recommended for publication by Editor Chao-Bo Yan upon evaluation of the Associate Editor and Reviewers' comments. This work was partially supported by DARPA award HR00112490426 and Sony Research Award. (Corresponding author: Chen Wang.)}
\thanks{\textsuperscript{1}Qiwei Du, Yi Du, Shaoshu Su, Taimeng Fu, Zitong Zhan, Zhipeng Zhao, and Chen Wang are with the Spatial AI \& Robotics Lab, Department of Computer Science and Engineering, University at Buffalo, Buffalo, NY 14260, USA. Contact: \url{https://sairlab.org}.}%
\thanks{\textsuperscript{2}Bowen Li is with Carnegie Mellon University, Pittsburgh, PA 15213, USA.}%
\thanks{Source code and videos are available at \url{https://sairlab.org/flax}.}
\thanks{Digital Object Identifier (DOI): see top of this page.}
}

\begin{document}

\maketitle

\markboth{IEEE Robotics and Automation Letters. Preprint Version. Accepted Janurary, 2026}
{Du \MakeLowercase{\textit{et al.}}: Fast Task Planning with Neuro-Symbolic Relaxation}

\begin{abstract}
Real-world task planning requires long-horizon reasoning over large sets of objects with complex relationships and attributes, leading to a combinatorial explosion for classical symbolic planners.
To prune the search space, recent methods prioritize searching on a simplified task only containing a few ``important" objects predicted by a neural network.
However, such a simple neuro-symbolic (NeSy) integration risks omitting critical objects and wasting resources on unsolvable simplified tasks.
To enable Fast and reliable planning, we introduce a NeSy relaxation strategy (\shortname), combining neural importance prediction with symbolic expansion.
Specifically, we first learn a graph neural network to predict object importance to create a simplified task and solve it with a symbolic planner.
Then, we solve a rule-relaxed task to obtain a quick rough plan, and reintegrate all referenced objects into the simplified task to recover any overlooked but essential elements.
Finally, we apply complementary rules to refine the updated task, keeping it both reliable and compact.
Extensive experiments are conducted on both synthetic and real-world maze navigation benchmarks where a robot must traverse through a maze and interact with movable obstacles.
The results show that \shortname boosts the average success rate by 20.82\% and cuts mean wall-clock planning time by 17.65\% compared with the state-of-the-art NeSy baseline.
We expect that \shortname offers a practical path toward fast, scalable, long-horizon task planning in complex environments.

\end{abstract}


\section{Introduction}
\IEEEPARstart{T}{ask} planning, which involves determining \emph{what to do} and \emph{in what order}, is a critical capability in many robotic applications, such as warehouse operations~\cite{leet2023task}, domestic service~\cite{wang2020home}, planetary exploration~\cite{schuster2019towards}, and disaster response~\cite{osooli2024multi}.

Classical task planners like Fast Downward~\cite{helmert2006fast} offer formal guarantees on problems encoded in Planning Domain Definition Language (PDDL)~\cite{edelkamp2004pddl2} but scale poorly with the number of objects and relations~\cite{li2024logicity}.
For instance, consider a task where a robot must navigate through a maze-like environment while manipulating movable obstacles, executing primitive actions like move, push, pickup, and place.
Each additional operator exponentially increases the branching factor of the action space, inducing a combinatorial explosion in the discrete task planning graph, which can encompass millions of reachable states prior to incorporating continuous motion planning constraints.
For example, on a $15\!\times\!15$ maze, a state-of-the-art classical planner can spend several minutes expanding useless nodes even when equipped with advanced heuristics such as $h$\textsuperscript{FF}~\cite{hoffmann2001ff} and LM-cut~\cite{helmert2009landmarks}.
This is an unacceptable delay for many embedded systems with strict real-time requirements.

\begin{figure}[t]
  \centering
    \includegraphics[width=1\linewidth]{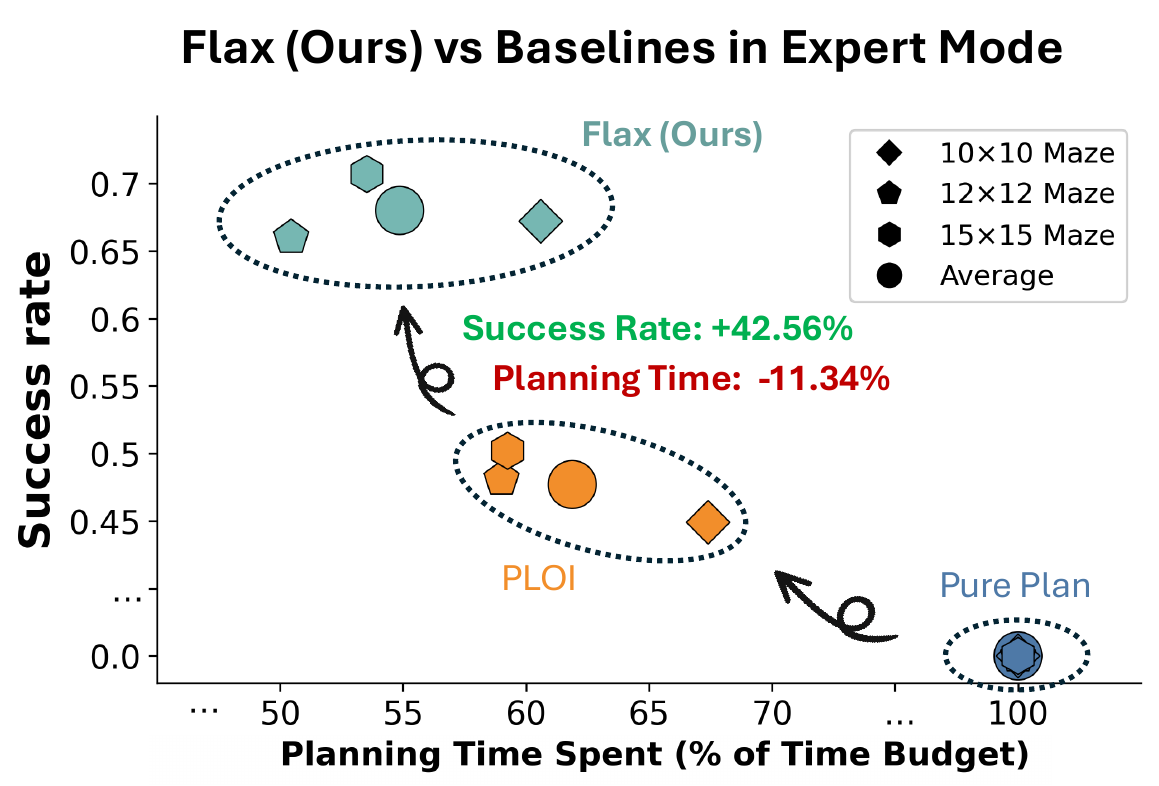}
    \caption{\small \textbf{\shortname achieves high success rates and fast planning on \textit{MazeNamo} tasks.} 
    The chart shows that \shortname significantly outperforms \emph{Fast Downward (Pure Plan)}~\cite{helmert2006fast} and \emph{PLOI}~\cite{silver2021planning} in both success rate and planning speed for \textit{expert-level} tasks.}
  \label{fig:teaser}
  \vspace{-10pt}
\end{figure}

To address this, two branches of research have emerged. 
The first learns an efficient neural policy via deep reinforcement learning (DRL) \cite{mnih2015human}.
Despite the promising results, most algorithms remain highly sample-inefficient~\cite{torrado2018deep} and generalize poorly on novel states~\cite{li2024logicity}. 
For PSPACE-hard planning domains \cite{bylander1994computational}, the size of policy grows exponentially with the number of objects~\cite{vega2020task, mao2023planning}, rendering the policies too large for memory-constrained hardware and infeasible for most fixed-size neural networks.
The second direction~\cite{silver2021planning, yang2022sequence, chen2024graph} learns a neural predictor that shrinks the search space for symbolic planners.

For instance, PLOI~\cite{silver2021planning} utilizes a Graph Neural Network (GNN) to filter out irrelevant objects, thus accelerating symbolic search. This approach necessitates a symbolic planner to iteratively verify task solvability as the filtering threshold decreases. However, in complex, long-horizon tasks with large search spaces, the planner often requires considerable time to determine unsolvability. Furthermore, due to the absence of correction mechanism, even minor mispredictions can exhaust the entire computational budget. This underscores our key insight: effective handling of complex planning tasks demands a lightweight correction mechanism to promptly detect and correct pruning errors, preventing unnecessary overhead.

To this end, we propose \textbf{\shortname}, a neuro-symbolic re\textbf{lax}ation strategy for \textbf{F}ast task planning that \emph{couples neural importance prediction with symbolic rough-plan enhancement and complementary rule expansion}. 
Specifically, we first employ a neural predictor with a short time budget to create an initial simplified task. 
If the task is not solved, we solve a relaxed version of the full task with domain-specific symbolic relaxation rules; every object involved by the rough plan is merged back to recover objects that the predictor may have missed. The symbolic search at this step does not aim to solve the full task, but rather to identify potentially important objects, so that pruning errors can be fixed without much effort.
Finally, we apply symbolic complementary rules to restore more potential missing objects. 
This strategy offers a fast local consistency recovery after rough-plan enhancement and thus largely preserves overall success rate. 
Together, they combine the raw speed of neural filtering with the reliability of symbolic reasoning, producing simplified tasks that are far smaller than the original one yet significantly decrease the possibility of dead-end failures.

We evaluate \shortname on a new \textbf{Maze Navigation among movable obstacles (\textit{MazeNamo})} benchmark in MiniGrid~\cite{MinigridMiniworld23} (three map sizes and four difficulty levels), high-fidelity Isaac Sim~\cite{makoviychuk2021isaac} 
forklift-warehouse scenarios, and real-world trials on a Unitree Go2 with a D1 manipulator. Under identical time budgets, \shortname boosts the average success rates by 20.82\% and cuts mean wall-clock planning time by 17.65\% relative to the previous neuro-symbolic baseline, while classical planning alone struggles on most \textit{hard} and \textit{expert} tasks, as is shown in~\fref{fig:teaser}. These gains carry over to real-world deployment: the same model drives the quadruped to pick, place, and push obstacles in cluttered real-world scenes, closing the loop from abstract PDDL reasoning to kinodynamic execution with no additional tuning. 
We summarize our contributions as follows:
\begin{itemize}[noitemsep,topsep=0pt]
    \item \textbf{Framework:} We present \shortname, a neuro-symbolic task planner integrating neural importance prediction, symbolic rough-plan enhancement, and complementary rule expansion to boost scalability and reliability.
    \item \textbf{Symbolic Relaxation and Complementation:} We propose a domain-specific symbolic relaxation and an object recovery method to quickly correct neural prediction errors, mitigating planning dead-ends.
    \item \textbf{\textit{MazeNamo} Benchmark:} We introduce a comprehensive benchmark including MiniGrid, Isaac Sim forklift-warehouse simulations, and real-world scenarios, validating the robustness and scalability of our method.
\end{itemize}

\section{Related Work}
\label{sec:related}

\subsection{Neuro-Symbolic Task Planning for Robots}
Task planning produces discrete action sequences that achieve high-level goals in robotics~\cite{garrett2020pddlstream}, logistics~\cite{pedersen2015automated}, and games~\cite{silver2016mastering}. Classical forward-search planners such as Fast Downward~\cite{helmert2006fast} offer completeness and optimality guarantees, yet direct applicability to robotics is limited due to 
(1) They usually assume a fully discrete, factored state space and 
(2) The search complexity grows exponentially with the number of objects~\cite{li2024logicity}. 
To overcome these limitations, neuro-symbolic approaches integrate learned neural components into the planning loop.
One stream of work leverages neural networks to lift raw states into predicates~\cite{li2023embodied,silver2023predicate,hansen2022bisimulation,li2025bilevel} and raw actions into skills and operators~\cite{chitnis2022learning,silver2022learning,kumar2024practice}.
The resulting hybrid system makes long-horizon task-and-motion planning (TAMP) \textit{possible} on real robots~\cite{chitnis2022learning,silver2023predicate,kumar2023learning,li2025bilevel}.
A complementary line assumes the planning model is given and fixed, while uses neural networks to make planning more \textit{efficient}. Early systems learned heuristics or value functions to guide best-first search~\cite{arfaee2011learning,silver2016mastering}, while more recent methods predict object-level relevance with graph neural networks (GNNs)~\cite{silver2021planning,chen2024graph}.
Our work falls within the second stream: assuming a given symbolic domain and low-level skills, we focus on accelerating planning without modifying the predicates or operators.

\begin{figure*}[thpb]
  \centering
  \includegraphics[width=\textwidth]{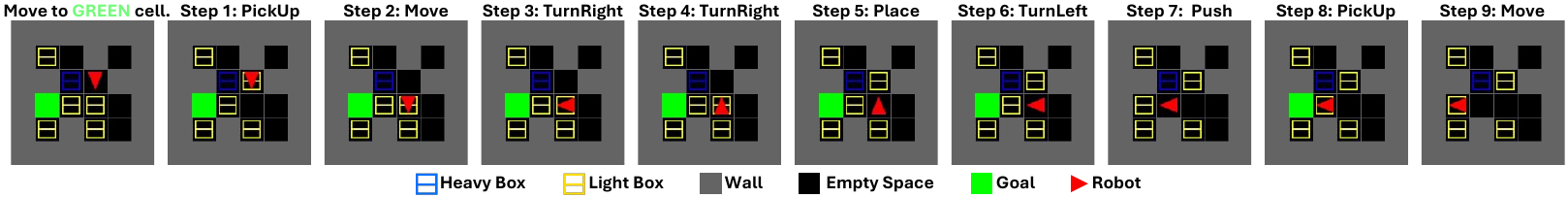}
  \caption{A typical maze navigation instance and automatically synthesized plan. \textit{Heavy boxes (push-only):} blue squares; \textit{Light boxes (push/pick-and-place):} yellow squares; \textit{Walls:} grey squares (cannot be manipulated); \textit{Empty space:} black squares; \textit{Goal:} green cell; \textit{Robot:} red triangle. Each snapshot shows the state immediately after the labelled operator executes.}
  \vspace{-10pt}
  \label{fig:example_preliminary}
\end{figure*}

\subsection{Learning for Efficient Task Planning}
\label{subsec:learn_eff}
Learning-based accelerations of task planning algorithms can be broadly categorized into three principal families.
\myparagraph{Learning generalized policies}
Policy-guided planning aims to learn feedforward program-like policies that can solve an entire family of problem instances.
Search-based methods tackle this in two steps: they first discover symbolic state features and then synthesize program policies over those features. 
PG3 exemplifies this approach by evaluating each candidate lifted policy through direct execution during training \cite{yang2022pg3}. Extending the idea to task-and-motion planning, Curtis \textit{et al.} automatically derive state and action abstractions that seamlessly couple discrete and geometric reasoning \cite{curtis2022discovering}. The same framework has recently been generalized to non-deterministic domains \cite{hofmann2024learning}.
Complementarily, deep reinforcement-learning approaches train Q-functions that are expressed with relational abstractions.
These abstractions enable zero-shot transfer to novel object configurations~\cite{karia2022relational}.

\myparagraph{Learning plan feasibility or informative subgoals}
Instead of producing full policies, another strand of work predicts whether partial plans are feasible or which subgoals to pursue, thus pruning unpromising branches during planning. Examples include sequence-based feasibility predictors for TAMP~\cite{yang2022sequence} and hierarchical RL that dynamically constructs landmark graphs to propose subgoals~\cite{zhang2023balancing}.

\myparagraph{Learning heuristics and relaxations}
A third family learns heuristic estimates or lightweight relaxations that bias classical search. Learned admissible heuristics~\cite{arfaee2011learning} and value-guided graph search~\cite{silver2016mastering} fall into this category. Related works also do graph learning for heuristics in discrete and numeric domains~\cite{silver2021planning,chen2024graph}. Our approach also falls into this category: we apply a domain-specific relaxation that yields a rough plan retaining only objects required for optimality, significantly reducing the grounded planning problem size.

\section{Preliminaries}
We define the basic concepts and introduce the classical 
\textbf{PDDL problem} \cite{fikes1971strips}, which is represented as a tuple:
\begin{equation}
\Pi = \langle \mathcal D,\;\tau  \rangle ,
\end{equation}
where $\mathcal D$ and $\tau$ denote a \textit{domain} and a \textit{task instance}.

\noindent A \textbf{Domain} $\mathcal D$ describes lifted symbols that are shared across all concrete planning tasks. Formally it is defined as:
\begin{equation}
  \mathcal D=\langle \mathcal P,\;\mathcal A \rangle ,
\end{equation}
where $\mathcal P$ is a finite set of \emph{lifted predicates} $p\in \mathcal P$, each defined over a list of typed variables; $\mathcal A$ is a finite set of \emph{lifted actions} where each action $\alpha\in \mathcal A$ is defined as a quadruple $\alpha = \langle \mathtt{pars}(\alpha),\; \mathtt{pre}(\alpha),\; \mathtt{add}(\alpha),\; \mathtt{del}(\alpha) \rangle $, $\mathtt{pars}(\alpha)$.
Here, $\mathtt{pars}(\alpha)$ is a list of typed variables, and $\mathtt{pre}(\alpha)$, $\mathtt{add}(\alpha)$, and $\mathtt{del}(\alpha)$ are sets of lifted predicates, with all free variables drawn from $\mathtt{pars}(\alpha)$. Specifically, if a set of concrete objects is specified for a particular task, grounding the predicates yields the corresponding state space.

\myparagraph{Task}$\tau$ is defined as
\begin{equation}
  \tau=\langle \mathcal O,\;I,\;G \rangle,
\end{equation}
where $\mathcal O$ is a finite set of \emph{typed objects}. Following \cite{silver2021planning}, we also treat object types as unary predicates.
We denote $P(\mathcal O)$ as the set obtained by grounding all lifted predicates with objects from $\mathcal O$, and a \textit{state} is any subset $s\!\subseteq\mathcal  \!P(\mathcal O)$. Thus, $I\!\subseteq\mathcal \!P(\mathcal O)$ is the \textit{initial state}, and
$G\!\subseteq\mathcal \!P(\mathcal O)$ is a set of \textit{grounded goal atoms} that must be \textit{true} after plan execution. 

\myparagraph{Plans}
A \emph{grounded action} $a$ is obtained by replacing the variable parameters of a lifted action $\alpha$ with appropriately typed objects from $\mathcal O$. 
A finite sequence of grounded actions
$
  \langle a_1,a_2,\ldots,a_{H}\rangle
$
is a \textit{plan} for task $\tau$.


\section{Methodology}

\subsection{MazeNamo Problem Setup}\label{sec:domain}
In this work, we define a challenging maze navigation problem among movable obstacles, where a robot must navigate to a goal position while manipulating both \emph{heavy} and \emph{light} boxes. Some general rules and constraints are:
\begin{itemize}
    \item \textit{Heavy} boxes can only be pushed on the floor. 
    \item \textit{Light} boxes can be pushed or picked up by the robot. 
    \item A \textit{light} box can be placed either on the floor or on top of a heavy box (but not on another light box).
    \item The \textit{robot} may manipulate only one box at a time (no simultaneous multi-box pushes or multi-box pickups).
    \item \textit{Walls} cannot be manipulated. 
\end{itemize}

\myparagraph{PDDL Domain} is composed of the following elements:
\begin{itemize}
\item Types: \texttt{\small ?r - robot}, \texttt{\small ?o - obstacle}, \texttt{\small ?p - pos}.
\item Key predicates: robot/obstacle position (\texttt{\small rAt(?r,?p)}, \texttt{\small oAt(?o,?p)}), orientation (\texttt{\small dirIsLeft(?r)}, \dots), adjacency (\texttt{\small upTo(?p,?p)}, \texttt{\small leftTo(?p,?p)}, \dots), obstacle category (\texttt{\small isHeavy(?o)}, \texttt{\small isLight(?o)}), obstacle status (\texttt{\small onGround(?o)}, \texttt{\small upon(?o,?o)}, \texttt{\small clear(?o)}), and grasp state (\texttt{\small handempty(?r)}, \texttt{\small holding(?r,?o)}).
\item Action classes: Each abstract capability (\texttt{\small Turn}, \texttt{\small Move}, \texttt{\small Push}, \texttt{\small PickUp}, and \texttt{\small Place}) is realized by several concrete operators. For example, \texttt{\small Place} is split into ``on ground'' vs. ``on obstacles'' variants, each expanding to four orientation‑specific actions. 
\end{itemize}

\myparagraph{Goal}
Each instance specifies a single goal literal \texttt{rAt(robot, p\textsubscript{goal})} (robot reaching a specific position).

\myparagraph{Challenges and Illustrative Example}
\fref{fig:example_preliminary} shows a typical maze. A heavy box (blue) blocks the path to the green goal, bordered by two stacked light boxes (yellow) and walls (grey). The robot must: (i) pick up the front light box, (ii) relocate it, (iii) turn left and push, (iv) lift the second light box, and (v) move into the now-clear goal cell, yielding the 9-step plan in \fref{fig:example_preliminary}. Compared to \textit{Sokoban}~\cite{culberson1997sokoban}, \textit{MazeNamo} adds two box categories, stacking, pick-and-place, and orientation constraints, largely
expanding the search space and branching factor. Standard heuristic planners thus face harder instances, and neural relaxations often fail on long horizons where search cannot easily recover from mispredictions. On the $15\!\times\!15$ expert-level mazes, the classical planner \textit{Fast Downward} takes over 100 seconds on average to find a \textit{satisficing} plan, which is too slow for real-world applications with strict time budgets.

\begin{figure*}[thpb]
  \centering
  \includegraphics[width=0.95\textwidth]{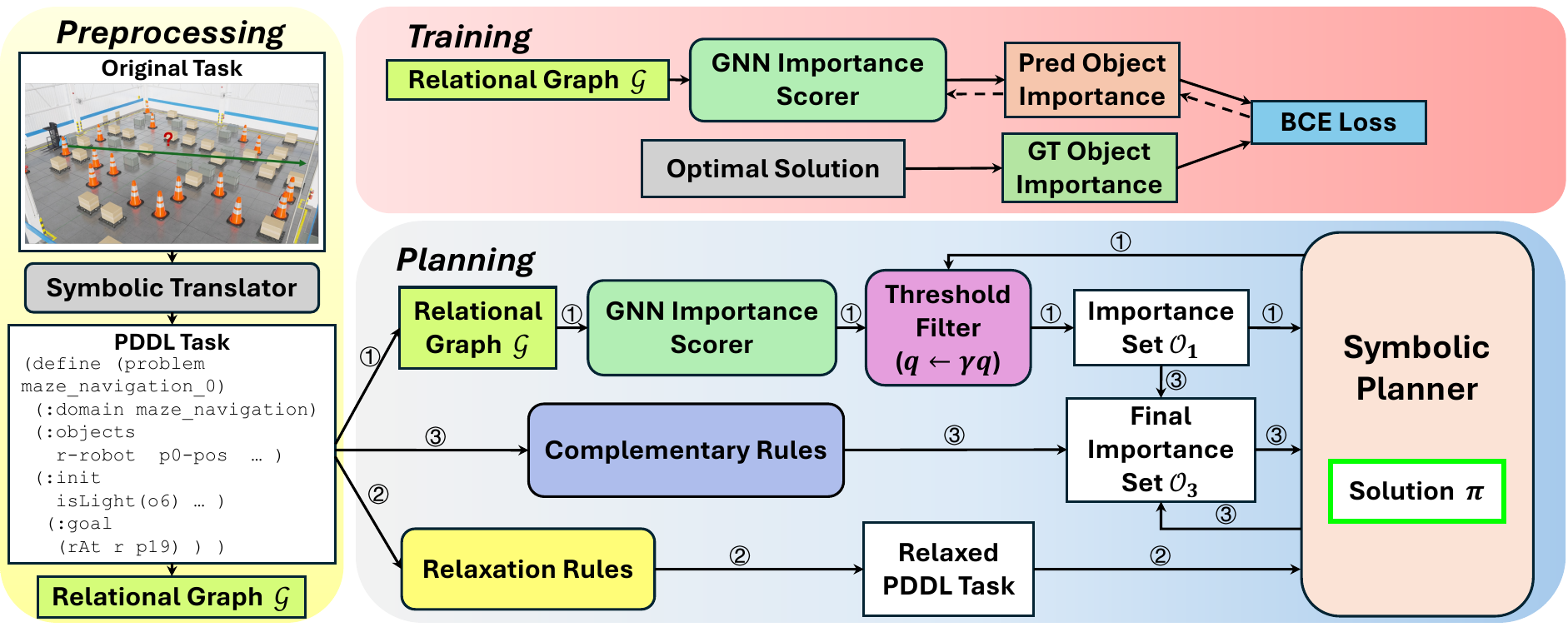}
  \caption{\small \textbf{\shortname\ pipeline.}
  \textbf{Pre-processing:} convert the task to PDDL and encode it as a relational graph \(\mathcal{G}\).
  \textbf{Training:} a symbolic planner provides an optimal plan; objects in the plan are labeled 1 as important, others 0. A GNN learns these labels from \(\mathcal{G}\) with BCE loss.
  \textbf{Planning (budget \(B\))}: planning runs in three steps.
  \emph{Step 1:} keep objects with importance \(>q\) to form \(\tau_{\mathcal{O}_1}\); lower \(q\) until a plan \(\mu\) is found or time \(\Delta t_1\) expires.
  \emph{Step 2:} if Step 1 fails, solve a rule-relaxed task within \(\Delta t_2\) to obtain rough plan \(\mu_r\); merge its objects with \(\mathcal{O}_1\) to get \(\mathcal{O}_2\).
  \emph{Step 3:} lightweight rules restore consistency, yielding \(\mathcal{O}_3\). Remaining time plans on \(\tau_{\mathcal{O}_3}\); the plan is then validated on the original task.}
  \label{fig:pipeline}
\end{figure*}

\subsection{Problem Formulation}\label{sec:problem_formulation}
We study the following problem: Training a neural network to predict accurate importance scores for all the objects in a PDDL task.
This can be formulated as a bilevel optimization \cite{wang2025imperative}, where the lower level minimizes the total action cost using a symbolic planner and produces an \emph{optimal} plan, and the upper level minimizes the loss between the predicted importance scores from a neural network and the ground truth importance scores derived from the \emph{optimal} plan:
\begin{subequations}
\label{eq:bilevel}
\begin{align}
\min_{\bm{\theta}}\;&\;\mathcal U\bigl(s(\bm{\theta},\tau),\bm{\mu}^*\bigr),\\
  \text{s.\,t. } & \quad
      \bm{\mu}^* = \arg\min_{\bm{\mu}}\;
      \mathcal L(\bm{\mu};\tau),\label{eq:low-opt}
\end{align}
\end{subequations} where $\mathcal U$ is the upper-level objective, $s$ is the neural importance predictor with parameters $\bm{\theta}$, $\tau$ is the PDDL task to solve, $\bm{\mu}$ is the action sequence produced by the symbolic planner, and $\mathcal L$ is the lower-level objective.

\myparagraph{Upper-level Objective}
The neural importance predictor $s_{\theta}$ assigns a score $s_{\theta}(o)\in[0,1]$ to every object node $o$ in the relational graph induced by $\tau$ similar to \cite{silver2021planning}. 
Given an \emph{optimal} plan $\mu^*$ for task $\tau$, we assign each node a binary label
\[
  \ell(o)=
  \begin{cases}
     1 & \text{if } o \text{ appears in }\mathtt{pars}(a)\text{ for any }
         a \in \mu^{\star},\\[2pt]
     0 & \text{otherwise.}
  \end{cases}
\]
The neural parameters are learned by minimizing the lower-level objective, which is a binary cross-entropy (BCE) loss
\begin{equation}
\begin{split}
  \mathcal{U}
  \!=\!\frac{-1}{|V|} \sum_{o \in V} \Bigl[
    \ell(o)\log s(o)
    \!+\!\bigl(1\!-\!\ell(o)\bigr)\log\bigl(1\!-\!s(o)\bigr)
  \Bigr],
\end{split}
\label{eq:bce}
\end{equation}
where $V$ denotes the object set of $\tau$.

\myparagraph{Lower-level Objective}
The lower-level problem solves the PDDL task $\tau$ using a symbolic planner, which searches over all feasible action sequences and returns an \emph{optimal} plan $\mu^{\star}$ under the domain's plan-cost metric.

\subsection{Training}
The GNN-based importance predictor $s_{\theta}$ is trained using supervision from optimal demonstrations. Each PDDL task $\tau$ is compiled into a relational graph $\mathcal G_\tau = (\mathcal V_\tau, \mathcal E_\tau)$, where nodes represent objects $o \in \mathcal O_\tau$ with feature vectors that include a one-hot type encoding, a multi-hot mask of true unary predicates in the initial state, and a goal-indicator bit for each unary predicate. Edges are created for each ordered pair of distinct objects ($o_i, o_j$), with edge features derived from the binary predicates $\mathcal P^{(2)}$, including Boolean indicators for their truth value in the initial state and whether they are required in the goal. Objects are labeled as \emph{important} ($\ell = 1$) if they appear in the grounded preconditions or effects of any action in the optimal plan; objects explicitly mentioned in the goal are always labeled as important. The model uses a 3-layer GNN architecture with 16-dimensional hidden layers, ReLU activation, and layer normalization, alternating between edge-to-node and node-to-edge message passing for 3 iterations. Final node representations are mapped to importance scores via a linear layer followed by a sigmoid function. The model is trained using the binary cross-entropy loss defined in \eref{eq:bce}. While this training process mirrors \cite{silver2021planning}, we emphasize a key difference: in our framework, the GNN’s predictions are used only as a first step. Since GNN-based importance scoring is not always reliable, especially under limited supervision, downstream symbolic planners may get stuck if critical objects are misclassified. This motivates our framework, which augments learned importance scores with rough-plan recovery and symbolic consistency checks to ensure robustness at planning.

\begin{figure*}[thpb]
  \centering
  \includegraphics[width=\textwidth]{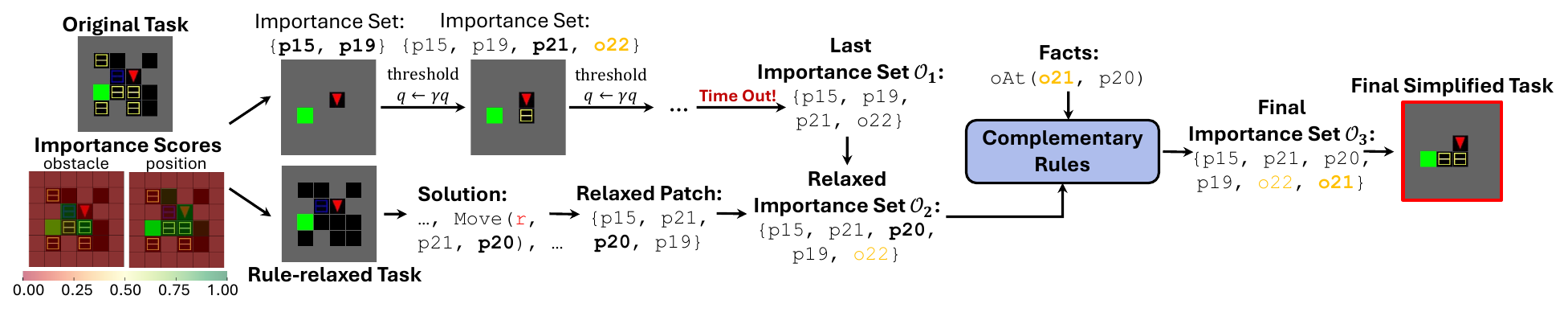}
  \caption{\textbf{Importance set progression:} pruning in Step 1 keeps $\{p_{15},p_{19}\}$, then lowering $\tau$ adds $\{p_{21}, o_{22}\}$; the rough plan injects $p_{20}$ into $\mathcal{O}_{1}$ to form $\mathcal{O}_{2}$; symbolic complementation adds $o_{21}$, yielding $\mathcal{O}_{3}$ that defines the solvable simplified task.}
  \vspace{-10pt}
  \label{fig:example_pipeline}
\end{figure*}

\subsection{Planning}
To solve the problem \eqref{eq:bilevel}, we introduce \textbf{\shortname}, a neuro-symbolic relaxation framework designed for fast task planning. It combines neural importance prediction with symbolic rough-plan refinement and rule-based expansion, progressively narrowing down the search space while retaining solvability.
The pipeline consisting of three planning steps is illustrated in \fref{fig:pipeline}.
To aid understanding, we use the example from \fref{fig:example_pipeline} throughout the explanation of each step.

\myparagraph{Planning Step 1 (Importance‑guided Pruning)}
In the planning stage, we first use the learned GNN importance predictor to assign a \emph{score} \(s(o)\!\in\![0,1]\) to every object \(o\). Starting from a high threshold \(q\!\leftarrow\! q_{\max}\), we keep only the objects with \(s(o)\ge q\), translate the resulting state into a \emph{simplified task}, and send it into a symbolic planner. If a solution is returned and verified to be valid on the original problem, the problem is solved; otherwise we gradually lower~\(q\) until a time budget is exhausted. 
This pruning strategy offers notable computational advantages by reducing the complexity of the planning problem. But unlike \cite{silver2021planning}, which allocates the entire planning budget to iterative pruning, we dedicate only a small fraction of the total time to this step. 
Our insight is that aggressive pruning quickly reveals a compact, expressive object subset that supports further refinement.
Threshold is initialized to $q_{\max}\!=\!0.81$ with a decay factor $\gamma\!=\!0.9$. After each failure, we set $q\!\leftarrow\!\gamma q$ and rebuild the simplified task $\tau_{\mathcal O}$ with $\mathcal O\!=\!\{o\!\mid \!s(o)\!\ge\!q\}$, yielding a bootstrap object set \(\mathcal O_1\).

\myparagraph{Planning Step 2 (Rough-plan Enhancement)}\;
\label{sec:step2}
Step 1 may exclude critical objects, rendering the simplified task unsolvable. To recover important objects efficiently, we solve a \emph{rule‑relaxed} version of the original task. The design of the relaxation rules follows a simple guideline: \textbf{\emph{making the problem substantially easier while still preserving a rough structural resemblance to the original task.}} Because the relaxed task is much simpler, a \emph{rough plan} \(\mu_r\) can be obtained significantly faster than the original problem. The key insight behind this step is leveraging domain-specific relaxation rules to quickly identify objects likely essential for solving the original problem from a symbolic perspective. We collect the set of objects involved in \(\mu_r\) (\(\mathcal O_r=\{o\mid o\text{ appears in }\mu_r\}\)) and merge it with \(\mathcal O_1\) to obtain \(\mathcal O_2=\mathcal O_1\cup\mathcal O_r\). In the \emph{MazeNamo} domain, \emph{Relaxation Rule} is to \emph{remove all light boxes}. This can be expressed as a first-order logic rule:
\begin{align*}\small 
& \forall \texttt{o}\,\forall \texttt{p}\;\Bigl(\texttt{isObstacle(o)}\land \texttt{isPos(p)}\land \texttt{isMovable(o)}\\
& \qquad\quad \land\texttt{isLight(o)}\land \texttt{oAt(o,p)}\\
& \;\; \rightarrow \;\;
\neg \texttt{isObstacle(o)} \land \neg \texttt{oAt(o,p)} \land \neg \texttt{isLight(o)}\\
& \qquad\quad \land \neg \texttt{isMovable(o)} \land
\texttt{isEmpty(p)}\Bigr).
\end{align*}

For example, in \fref{fig:example_pipeline}, after two iterations in Step 1, the simplified task remains unsolvable when the time budget of Step 1 expires. In the rule-relaxed task, the solution involves position \texttt{p20}, a critical object ignored in the importance set, whose absence contributed to the simplified task's unsolvability. Position \texttt{p20} is therefore added to the last importance set to form the relaxed importance set.

\begin{table*}[thpb]
    \centering
    \caption{Performance of classical Fast Downward (\textsc{Pure Plan}), PLOI, PLOI with complementary rule expansion (\textsc{Comp.}), PLOI with rough-plan enhancement (\textsc{Relax.}), and the proposed \shortname. ``SR" denotes Success Rate, which is higher-better; ``WPT" denotes weighted planning time, assigning the full time budget to failures, which is lower-better. The last two colums show relative gains of \shortname over PLOI. The final row reports the average SR and WPT usage rate across all tasks. Since time budgets differ by difficulty level, WPT values are normalized by their respective budgets and averaged as percentages.}
    \label{tab:minigrid}
    \small
    \setlength{\tabcolsep}{5pt}
    \renewcommand{\arraystretch}{1.05}
    \begin{tabular}{l c c c c c c c c c c c c}
        \toprule
        \multirow{2}{*}{\textbf{Task}} & \multicolumn{2}{c}{\textbf{\textsc{Pure Plan}}} & \multicolumn{2}{c}{\textbf{PLOI}} & \multicolumn{2}{c}{\textbf{PLOI+\textsc{Comp.}}} & \multicolumn{2}{c}{\textbf{PLOI+\textsc{Relax.}}} & \multicolumn{2}{c}{\textbf{\shortname (Ours)}} & \multicolumn{2}{c}{\textbf{Improv.}} \\
        & \makecell[c]{\textbf{SR}} & \makecell[c]{\textbf{WPT (s)}} & \makecell[c]{\textbf{SR}} & \makecell[c]{\textbf{WPT (s)}} & \makecell[c]{\textbf{SR}} & \makecell[c]{\textbf{WPT (s)}} & \makecell[c]{\textbf{SR}} & \makecell[c]{\textbf{WPT (s)}} & \makecell[c]{\textbf{SR}} & \makecell[c]{\textbf{WPT (s)}} & \makecell[c]{\textbf{SR}} & \makecell[c]{\textbf{WPT}} \\
        \midrule
        10 (Easy) & \underline{\textbf{1.000}} & \underline{\textbf{0.94}} & 0.929 & 1.37 & 0.921 & 1.29 & 0.880 & 1.18 & 0.959 & 1.08 & +3.23\% & -20.81\% \\
        10 (Medium) & \underline{\textbf{1.000}} & 1.99 & 0.892 & 1.40 & 0.915 & 1.36 & 0.874 & 1.24 & 0.943 & \underline{\textbf{1.21}} & +5.72\% & -13.32\% \\
        10 (Hard) & 0.750 & 4.27 & 0.732 & 2.15 & 0.816 & 2.07 & 0.711 & 2.08 & \underline{\textbf{0.872}} & \underline{\textbf{1.93}} & +19.13\% & -10.34\% \\
        10 (Expert) & 0.000 & 5.00 & 0.449 & 3.37 & 0.480 & 3.35 & 0.514 & 3.10 & \underline{\textbf{0.672}} & \underline{\textbf{3.03}} & +49.67\% & -10.12\% \\
        \midrule
        12 (Easy) & \underline{\textbf{1.000}} & 4.35 & 0.883 & 3.86 & 0.913 & 3.58 & 0.921 & 2.88 & 0.971 & \underline{\textbf{2.47}} & +9.97\% & -36.16\% \\
        12 (Medium) & \underline{\textbf{1.000}} & 8.86 & 0.830 & 4.96 & 0.901 & 4.47 & 0.880 & 3.94 & 0.967 & \underline{\textbf{3.19}} & +16.51\% & -35.69\% \\
        12 (Hard) & 0.480 & 18.85 & 0.537 & 10.66 & 0.619 & 10.16 & 0.659 & 8.73 & \underline{\textbf{0.809}} & \underline{\textbf{7.69}} & +50.65\% & -27.88\% \\
        12 (Expert) & 0.000 & 20.00 & 0.481 & 11.80 & 0.516 & 11.57 & 0.600 & \underline{\textbf{10.07}} & \underline{\textbf{0.660}} & 10.09 & +37.21\% & -14.47\% \\
        \midrule
        15 (Easy) & \underline{\textbf{1.000}} & 21.46 & 0.848 & 9.95 & 0.887 & 9.55 & 0.889 & 8.41 & 0.968 & \underline{\textbf{7.41}} & +14.15\% & -25.58\% \\
        15 (Medium) & 0.630 & 36.35 & 0.793 & 12.42 & 0.831 & 12.16 & 0.855 & 10.70 & \underline{\textbf{0.932}} & \underline{\textbf{10.57}} & +17.53\% & -14.89\% \\
        15 (Hard) & 0.080 & 39.78 & 0.721 & 14.68 & 0.800 & 14.03 & 0.810 & 12.62 & \underline{\textbf{0.927}} & \underline{\textbf{11.86}} & +28.57\% & -19.21\% \\
        15 (Expert) & 0.000 & 40.00 & 0.502 & 23.70 & 0.537 & 23.45 & 0.644 & 21.95 & \underline{\textbf{0.707}} & \underline{\textbf{21.41}} & +40.84\% & -9.67\% \\
        \midrule
        Average & 0.578 & 70.68\% & 0.716 & 39.49\%	 & 0.761 & 38.19\% & 0.770 & 34.52\% & \underline{\textbf{0.866}} & \underline{\textbf{32.52\%}} & +20.82\% & -17.65\% \\
        \bottomrule
    \end{tabular}
    \vspace{-10pt}
\end{table*}

\myparagraph{Planning Step 3 (Complementary Rule Expansion)}
\label{sec:step3}
Finally, we apply symbolic \emph{Complementary Rules} \(\textsc{Comp}(\cdot)\) that enforce local consistency by ensuring that objects' initial relationships are maintained in the final importance set. The guiding intuition here is that objects connected by fundamental spatial or relational constraints in the initial state should either both be included or both be excluded. Violating these local consistencies typically results in unsolvable tasks or inefficiencies. The object set \(\mathcal O_3=\textsc{Comp}(\mathcal O_2)\) defines the \emph{final simplified task}, which is forwarded to the symbolic planner. 
In the \emph{MazeNamo} domain, we use:
\begin{align*}
&\forall \texttt{o}\;\forall \texttt{p}\;(\texttt{o}\in \mathcal{O} \,\land\,\texttt{oAt(o,p)}\;\rightarrow\;\texttt{p}\in \mathcal{O}),\\
&\forall \texttt{o}\;\forall \texttt{p}\;(\texttt{p}\in \mathcal{O} \,\land\,\texttt{oAt(o,p)}\;\rightarrow\;\texttt{o}\in \mathcal{O}),
\end{align*}
where $\mathcal{O}$ represents the importance set. These \emph{Complementary Rules} enforce mutual inclusion of \texttt{o} and \texttt{p} in the final importance set whenever \texttt{oAt(o,p)} holds in the initial state. For example, in \fref{fig:example_pipeline}, since \texttt{p20} is in the importance set, \texttt{o21} should also be added into the importance set, otherwise the simplified task 
is still unsolvable.

By combining GNN-derived importance, rough plan from rule-based relaxation, and complementary rule expansion, our method recovers missing objects critical for plan validity, while keeping the simplified task size manageable.

\section{Experiments}
\subsection{Implementation Details}
\myparagraph{Benchmark}
We introduce \textit{MazeNamoEnv}, a MiniGrid environment that generates random maze navigation tasks on an \(n \times n\) grid. The outer boundary consists of walls, and each interior cell is independently assigned as a wall (20\%), heavy box (10\%), light box (15\%), or free space (55\%). Tasks are symbolically translated into \textit{typed} PDDL instances \(\tau\) mentioned in \sref{sec:domain}.
Each instance is solved in \emph{satisficing} mode and categorized into four difficulty levels (\emph{easy}, \emph{medium}, \emph{hard}, and \emph{expert}) based on time cost. Unsolvable, trivial (\(<\!0.1\)s), or excessively difficult instances are discarded.
For map sizes \(n\!\in\!\{10, 12, 15\}\), we retain 300, 200, 100, and 100 instances per difficulty level, respectively, with evaluation time limits of 5s, 20s, and 40s. Another 200 easy instances with \(n\!=\!10\) are used for training. 
All the \emph{optimal} plans are generated by \emph{Fast Downward} in \texttt{seq-opt-lmcut} mode. And all the \emph{satisficing} symbolic planners in planning are \emph{Fast Downward} in \texttt{LAMA-first} mode.
To demonstrate sim-to-real generalization, the policy is executed in (i) Isaac Sim with a forklift platform and (ii) the real world using a Unitree Go2 quadruped manipulator with a D455 camera on the gripper.

\begin{figure}[t]
  \centering
  \includegraphics[width=0.5\textwidth]{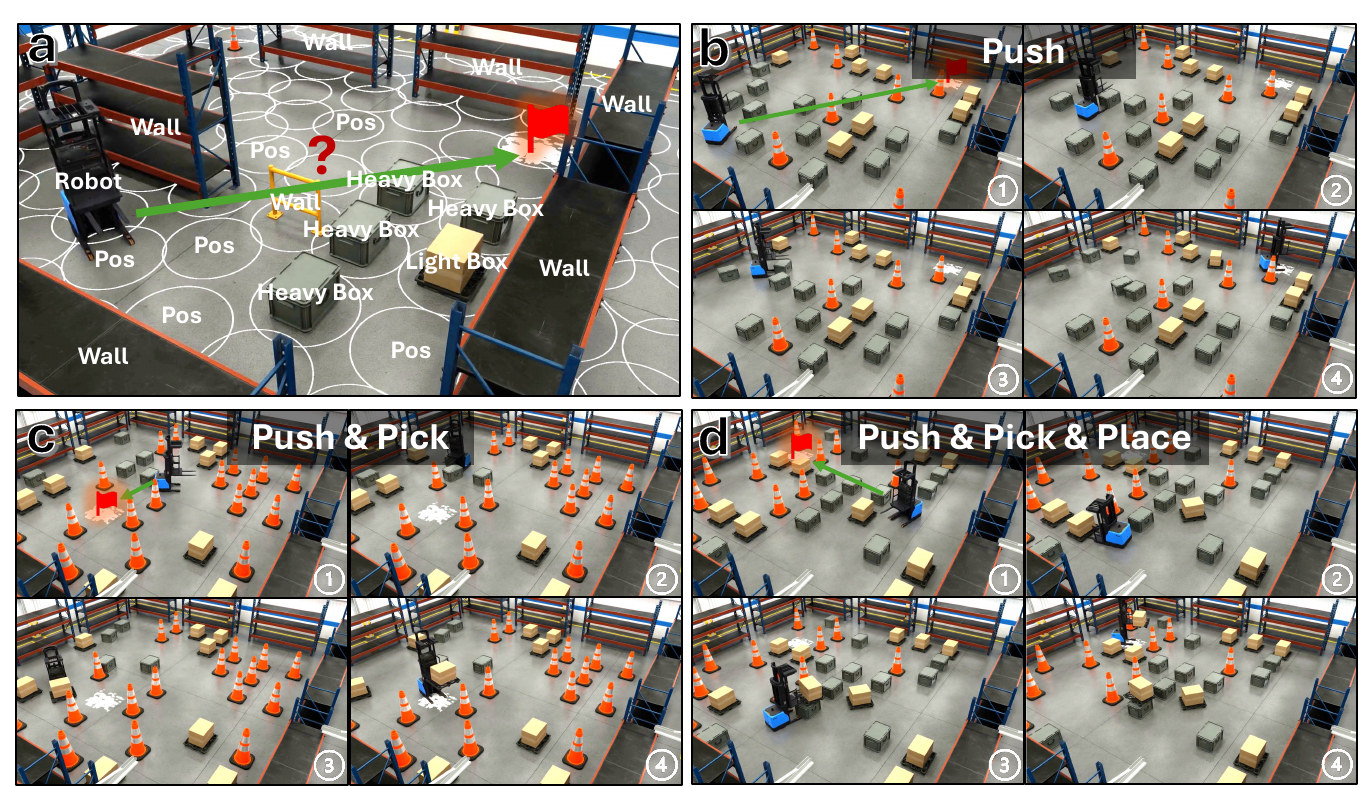}
  \caption{\textbf{IsaacSim Experiments.} (a) Scene abstraction in a robotics environment. Obstacles and positions are mapped to symbols. (b)(c)(d) IsaacSim tasks involving different actions.}
  \label{fig:isaacsim}
  \vspace{-10pt}
\end{figure}

\myparagraph{Evaluation Metrics}
We evaluate performance using two metrics:
\textit{Success Rate (SR)}, the proportion of test instances successfully solved within the time budget; and
\textit{Weighted Planning Time (WPT)}, the average time spent across all instances, where solved cases contribute their solver time and unsolved ones are counted as taking the full time budget.

\myparagraph{Adaptation to Robotics Environment}\;
We deploy the exact model trained in MiniGrid simulation on both a forklift in Isaac Sim and a Unitree Go2 quadruped equipped with a 6-DoF D1 manipulator \emph{without finetuning}.
First, a geometry-aware state abstraction maps each obstacle to a symbolic position based on its 
bounding box, transforming a 3D scene into a partial PDDL instance. Second, to
avoid free areas blocking valid path searching, we improves graph connectivity by sampling additional ``stepping-stone" positions using circle packing until saturation, as shown in \fref{fig:isaacsim} (a). Lastly, we translates high-level symbolic plans into platform-specific low-level skills, enabling deployment on diverse systems.

\begin{figure*}[thpb]
  \centering
  \includegraphics[width=\textwidth]{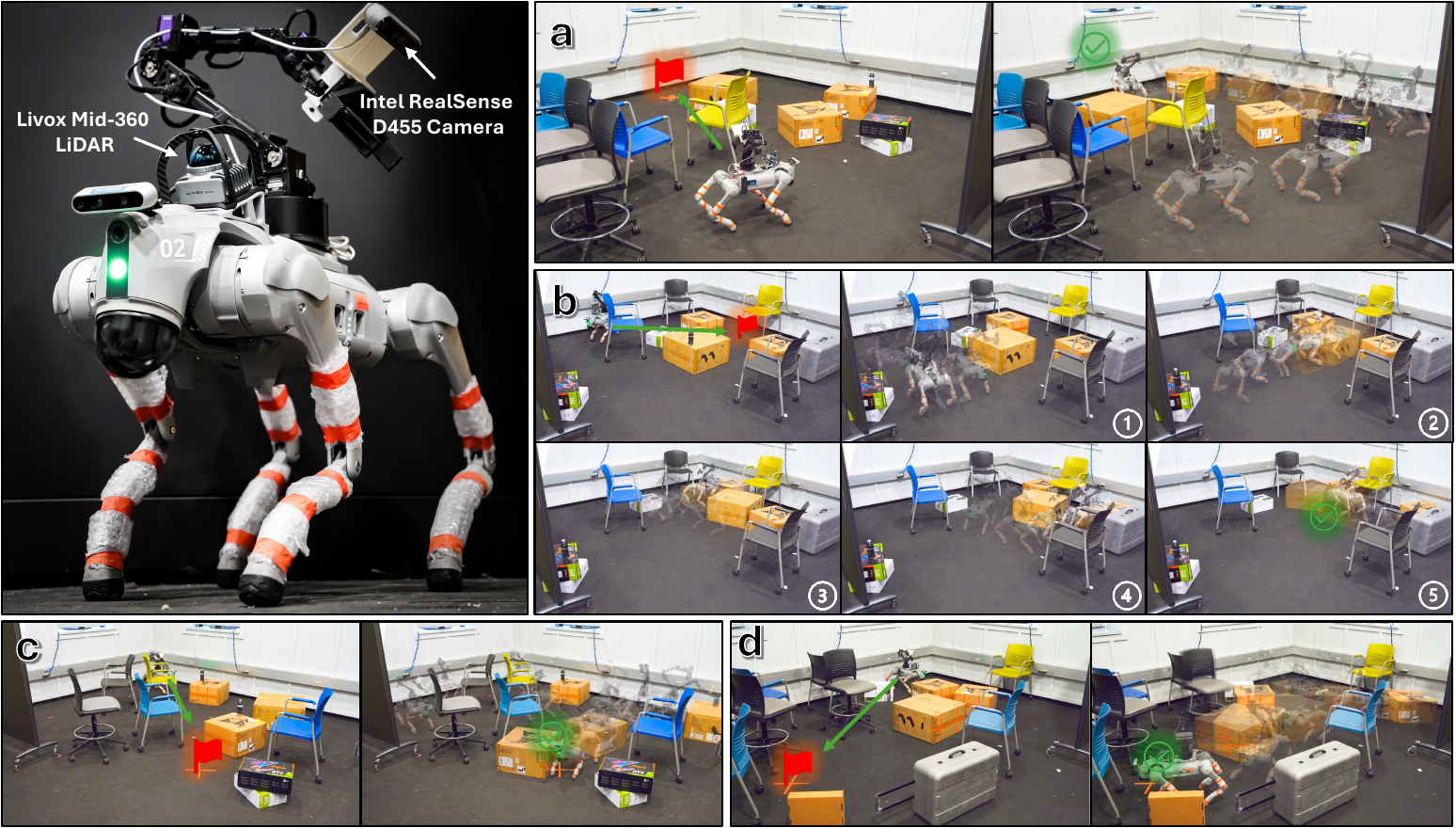}
  \caption{\textbf{Real-world Experiments.} A Unitree Go2 quadruped robot, equipped with a 6-DoF D1 arm, a Livox LiDAR, and an Intel RealSense camera (left), performs \textit{MazeNamo} tasks in real-world environments with varied layouts. In each task, the robot navigates toward the red flag marker while avoiding or manipulating obstacles. Semi-transparent overlays show the robot’s trajectory over time, illustrating the effectiveness of \shortname's plan execution in complex, real-world scenarios.}
  \vspace{-10pt}
  \label{fig:real_world}
\end{figure*}

\subsection{Baselines}
We compare \shortname against four baselines, each run with the same time budgets (5\,s, 20\,s, 40\,s for map sizes $n\in\{10,12,15\}$) and repeated for 10 random seeds.

\myparagraph{\textsc{Pure Plan}}\;
A baseline that plans directly on the complete PDDL problem instance, without relaxation. 

\myparagraph{\textsc{PLOI}}\;
Planning with Learned Object Importance in \cite{silver2021planning}.

\myparagraph{\textsc{PLOI}\,+\,\textsc{Comp.}}\;
A two-step variant that augments PLOI with complementary rules. Step~1 runs standard PLOI, and Step~2 expands the resulting simplified task using $\textsc{Comp}(\cdot)$.

\myparagraph{\textsc{PLOI}\,+\,\textsc{Relax.}}\;
Another two-step ablation that combines PLOI with only rule-relaxation heuristics, without $\textsc{Comp}(\cdot)$.

All baselines share the same GNN architecture and weights, ensuring observed differences are due to the object-selection strategy rather than the importance estimator.

\subsection{MiniGrid Experiment Results}\label{sec:sim-results}
\tref{tab:minigrid} summarizes performance on the \textit{MazeNamoEnv} benchmark across twelve test suites with three map sizes ($10\!\times\!10$, $12\!\times\!12$, $15\!\times\!15$) and four difficulty levels.

\myparagraph{\shortname Achieves Consistent and Significant Gains.}
\shortname ranks highest in both metrics on average (0.866 success rate using only 32.52\% of time budgets), indicating strong generalization across all the difficulty levels. Moreover, it outperforms PLOI using neural relaxation across all size and difficulty combinations, achieving an average +20.82\% improvement in SR and a 17.65\% reduction in WPT. 

\myparagraph{Harder Task Brings Greater SR Gain.}
The performance gap between \shortname and all the baselines widens with task difficulty. For instance, comparing to PLOI, SR gains are +49.67\% on $10\!\times\!10$ \emph{expert}, +50.65\% on $12\!\times\!12$ \emph{hard}, and +40.84\% on $15\!\times\!15$ \emph{expert}. This confirms \shortname's advantage in complex planning scenarios where other methods struggle.

\myparagraph{Symbolic Relaxation Reliably Fixes Pruning Errors.}
The limited improvements of \textsc{PLOI+Comp.} and \textsc{PLOI+Relax.} suggest that partial fixes are insufficient. In contrast, \shortname's symbolic relaxation and complementation provide a principled mechanism to detect and correct pruning-induced omissions, leading to superior robustness.

\myparagraph{Scalability with Environment Size.}
As map size increases from $10\!\times\!10$ to $15\!\times\!15$, many baselines degrade significantly, while \shortname maintains strong performance. For example, although only trained on the smallest maps, its success rate remains above 0.9 even on \emph{hard} settings in the largest maps, confirming its scalability to larger domains.

\subsection{Isaac Sim Evaluation} 
We evaluated \shortname's performance on a forklift platform within the Isaac Sim environment. 
The same model, trained solely on MiniGrid environments, was directly deployed to plan in Isaac Sim. 
The results in \fref{fig:isaacsim} showed that \shortname successfully generated valid plans including complex object rearrangements, such as pushing heavy boxes (grey plastic boxes) or lifting and placing light boxes (yellow cardboard boxes) to clear paths, showcasing \shortname's ability to generalize from symbolic operations to real robot motions in physics-based simulation scenes with collisions and constraints.

\subsection{Real-World Evaluation}
To further validate \shortname's practical applicability, as shown in \fref{fig:real_world}, we deployed the trained model on a Unitree Go2 quadruped robot equipped with a Livox Mid-360 LiDAR, a 6-DoF D1 manipulator, and an Intel RealSense D455 camera mounted near the gripper. 
We evaluated six real-world tasks, each executed three times.
These real-world trials encompassed subtasks such as picking up and placing bottles, pushing boxes, and navigating through cluttered spaces. For example, in \fref{fig:real_world} (b) (Task 4), where all the chairs are seen as unmovable obstacles, the robot must reach a goal position obstructed by a bottle and multiple boxes. Due to task constraints (two boxes cannot be pushed simultaneously and a box cannot be pushed while a bottle is on top), the robot follows a sequential strategy: \ding{172} it navigates to the center box, lifts the bottle, and relocates it to free the box; \ding{173} pushes the center box toward the goal to create clearance; \ding{174} moves the two left-side boxes to make space for the box currently occupying the goal; \ding{175} pushes the right-side box to access a position suitable for pushing the center box; and \ding{176} finally pushes the center box away and reaches the goal position to complete the task.
\tref{tab:real_world_results} reports the success rates. Our system achieved an average success rate of 67\%, despite significant real-world challenges including inaccurate object localization and imperfect physical interactions.
\begin{table}[h]
\centering
\caption{Real-world evaluation on Go2 with a D1 arm.}
\label{tab:real_world_results}
\begin{tabular}{c|c|c}
\toprule
\textbf{Task ID} & \textbf{Success Rate} & \textbf{Failure Causes} \\
\midrule
1 & 3/3 & --- (No failures) \\
2 & 2/3 & Box rotation causing path blockage \\
3 & 2/3 & Box rotation causing path blockage \\
4 & 1/3 & Box rotation causing path blockage \\
5 & 2/3 & Bottle grasping failure \\
6 & 2/3 & Robot stuck when turning in a corner \\
\midrule
\textbf{Average} & \textbf{67\%} & --- \\
\bottomrule
\end{tabular}
\end{table}
The most common failure mode (Tasks 2–4) arises from the Go2 robot pushing boxes using its head, which provides only a single-point contact. When the push is not perfectly centered, the box may unexpectedly rotate, blocking the intended path. Additional failures include occasional bottle grasping errors (Task 5) and the robot becoming stuck during tight corner turns (Task 6).
This experiment showcases a comprehensive integration of complex reasoning and low-level action execution, demonstrating the effectiveness of \shortname when planning in physical domains. 
Despite the inherent complexities of real-world deployment, including sensor noise, minor calibration errors, and non-grid layouts, \shortname consistently enabled the robot to achieve its high-level goals. Importantly, this was accomplished without any real-world retraining, using the same policy trained entirely in low-fidelity simulation. 
Source code and videos of these real-world demonstrations are available at \url{https://sairlab.org/flax/}.

\section{Conclusion}
To summarize, we presented \shortname, a three-step neuro-symbolic task-planning framework that combines learned importance prediction with symbolic relaxation and expansion. This approach accelerates planning while improving robustness against incomplete task abstractions. Experiments on the \textit{MazeNamo} benchmark show a 20.82\% increase in success rate and a 17.65\% reduction in planning time compared to previous methods. The framework successfully generalizes to real-world tasks without additional fine-tuning, demonstrating its practical applicability. 
While this work focuses on offline bi-level training and the runtime relaxation–complementation pipeline, the broader bi-level perspective suggests several promising directions, including online refinement of the importance predictor, self-supervised incorporation of plans generated during deployment, and tighter integration between symbolic relaxations and learned relational models. Extending \shortname to continuous/hybrid task-and-motion domains, as well as enabling automatic discovery of relaxation rules, also represents valuable future work.




\section*{Acknowledgement}

Any opinions, findings, conclusions, or recommendations expressed in this paper are those of the authors and do not necessarily reflect the views of DARPA or Sony.


\bibliographystyle{IEEEtranBST/IEEEtran}
\bibliography{IEEEtranBST/IEEEabrv,refs/references}

\end{document}